\documentclass[bst/sn-mathphys-ay]{sn-jnl}

\usepackage{graphicx}%
\usepackage{multirow}%
\usepackage{mathrsfs}%
\usepackage[title]{appendix}%
\usepackage{xcolor}%
\usepackage{textcomp}%
\usepackage{manyfoot}%
\usepackage{booktabs}%
\usepackage{algorithm}%
\usepackage{algorithmicx}%
\usepackage[noend]{algpseudocode}%
\usepackage{listings}%
\usepackage{hyperref}
\usepackage{url}
\usepackage[utf8]{inputenc} 
\usepackage[T1]{fontenc}    
\usepackage{booktabs}       
\usepackage{nicefrac}       
\usepackage{microtype}      
\usepackage{xcolor}         
\usepackage{wrapfig}
\usepackage{multirow}
\usepackage{enumitem}
\usepackage{longtable, lmodern}
\usepackage{caption}
\usepackage{algorithm}
\usepackage{subcaption}
\usepackage{adjustbox}
\usepackage{tabularx}
\usepackage{xspace}
\usepackage[numbers]{natbib}
\usepackage{array}
\usepackage{todonotes}

\usepackage{tikz}
\usetikzlibrary{shapes.geometric, arrows.meta, positioning}

\title[Article Title]{GNN-MultiFix: Addressing the pitfalls for GNNs for multi-label node classification}

\author{\fnm{Tianqi} \sur{Zhao}}\email{T.Zhao-1@tudelft.nl}

\author{\fnm{Megha} \sur{Khosla}}\email{m.khosla@tudelft.nl}
\affil{\orgdiv{Intelligent Systems Department}, \orgname{Delft University of Technology}, \orgaddress{\country{The Netherlands}}}

\newcommand{\mpara}[1]{\medskip\noindent{\bf #1}}
\newcommand{\blog}{\textsc{BlogCat}\xspace}
\newcommand{\yelp}{\textsc{Yelp}\xspace}
\newcommand{\pcg}{\textsc{PCG}\xspace}

\newcommand{\dblp}{\textsc{DBLP}\xspace}

\newcommand{\featqua}{\textsc{Synthetic1}\xspace}
\newcommand{\homolevel}{\textsc{Synthetic2}\xspace}
\newcommand{\gcn}{\textsc{Gcn}\xspace}
\newcommand{\mlp}{\textsc{Mlp}\xspace}
\newcommand{\gat}{\textsc{Gat}\xspace}
\newcommand{\deepwalk}{\textsc{DeepWalk}\xspace}
\newcommand{\majorityvote}{\textsc{MajorityVote}\xspace}
\newcommand{\graphsage}{\textsc{GraphSage}\xspace}
\newcommand{\hgcn}{\textsc{H2Gcn}\xspace}
\newcommand{\gcnlpa}{\textsc{GCN-LPA}\xspace}
\newcommand{\lanc}{\textsc{LANC}\xspace}
\newcommand{\lspe}{\textsc{GNN-LSPE}\xspace}
\newcommand{\idgnn}{\textsc{ID-GNN-Fast}\xspace}

\newcommand{\fsgnn}{\textsc{FSGNN}\xspace}

\newcommand{\mymodel}{\textsc{GNN-MultiFix}\xspace}
\newcommand{\mymodelmlp}{\textsc{GNN-MultiFix-MLP1}\xspace}
\newcommand{\mymodelMlp}{\textsc{GNN-MultiFix-MLP3}\xspace}
\newcommand{\mymodellin}{\textsc{GNN-MultiFix-Linear}\xspace}

\usepackage{amsmath}
\usepackage{amsfonts}
\usepackage{bm}
\usepackage{amsthm}

\def\eqref#1{equation~\ref{#1}}

\def\1{\bm{1}}

\DeclareMathAlphabet{\mathsfit}{\encodingdefault}{\sfdefault}{m}{sl}
\SetMathAlphabet{\mathsfit}{bold}{\encodingdefault}{\sfdefault}{bx}{n}

\newtheorem{definition}{Definition}

\newtheorem{lemma}{Lemma}
\newtheorem{theorem}{Theorem}

\begin{document}
\maketitle




\begin{abstract}

Graph neural networks (GNNs) have emerged as powerful models for learning representations of graph data showing state of the art results in various tasks. Nevertheless, the superiority of these methods is usually supported by either evaluating their performance on small subset of benchmark datasets or by reasoning about their expressive power in terms of certain graph isomorphism tests. In this paper we critically analyse both these aspects through a transductive setting for the task of node classification. First, we delve deeper into the case of multi-label node classification which offers a more realistic scenario and has been ignored in most of the related works. Through analysing the training dynamics for GNN methods we highlight the failure of GNNs to learn over multi-label graph datasets even for the case of abundant training data. Second, we show that specifically for transductive node classification, even the most expressive GNN may fail to learn in absence of node attributes and without using explicit label information as input. To overcome this deficit, we propose a straightforward approach, referred to as \mymodel, that integrates the feature, label, and positional information of a node. \mymodel demonstrates significant improvement across all the multi-label datasets. We release our code at {\url{https://anonymous.4open.science/r/Graph-MultiFix-4121}}.

\end{abstract}




\section{{Introduction}}
\label{sec:introduction}

Graph neural networks (GNNs) have become highly effective models for representing graph-structured data, delivering state-of-the-art performance in various tasks. One such prototypical task is the node classification task, which involves predicting node labels based on the graph structure and a partially labeled node set. While contemporary works have demonstrated significant improvements in multi-class node classification, where each node has at most one label, a recent work \citep{zhao2024multilabel} highlights the limitations of current approaches in addressing multi-label node classification, where each node can have multiple labels.

Specifically, classical GNNs \citep{kipf2016semi, velickovic2018graph, DBLP:journals/corr/HamiltonYL17} show superior performance on multi-class graph datasets with high label homophily, where connected nodes typically share the same labels. However, in multi-label datasets, a node usually shares only a small fraction of common labels with its neighbors, resulting in low label homophily per edge. Consequently, the aggregated information from the local neighborhood of each node contains information about diverse characteristics of the nodes in its local neighborhoods, making it challenging for the GNNs to differentiate the most informative part to infer the labels of the ego nodes. 

While \citep{zhao2024multilabel} showed that GNNs exploiting higher order neighborhoods are equally insufficient for tackling multi-label node, we provide an even more surprising result. In particular, we demonstrate that GNNs are significantly outperformed by a simple \majorityvote baseline, which relies solely on the label information of immediate neighbors from the training set to determine the label distribution of test nodes.


Taking a step further, we analyse the training dynamics of  representative methods for multi-label node classification. We validate the insufficiency of GNNs to learn over multi-label datasets even in case of abundant training data. 
\begin{wrapfigure}{r}{0.5\textwidth}
\vspace{-7mm}
\includegraphics[width=1.0\linewidth]{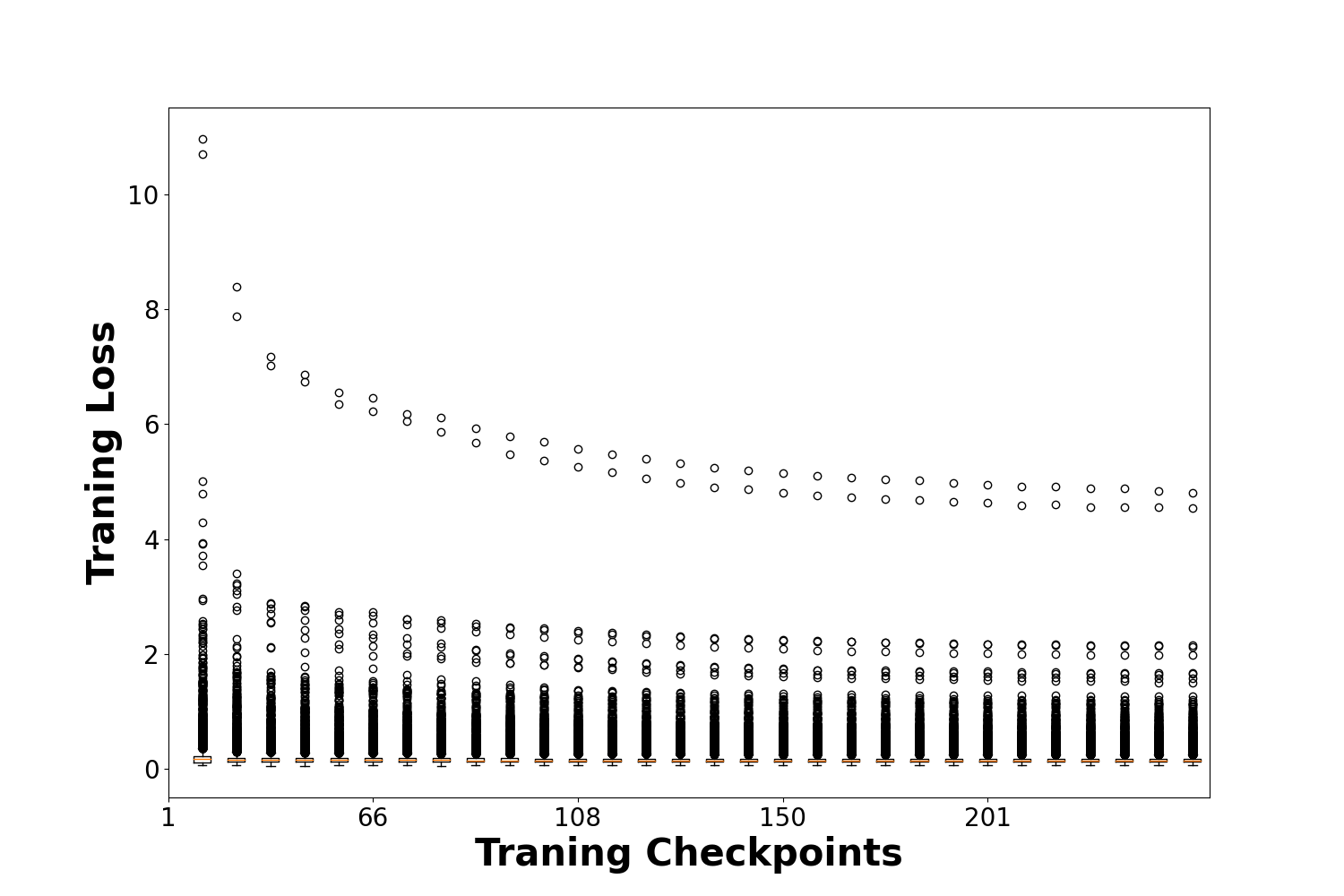} 
\caption{Training dynamics when \gcn is trained using all nodes of \blog}
\label{fig:trainingdynamics}
\vspace{-10mm}
\end{wrapfigure}
As an example in Figure \ref{fig:trainingdynamics} we plot box plots corresponding to training losses per node when all nodes of the multi-label dataset \blog \citep{Blogcatalog} are used for training \gcn \cite{kipf2016semi}. We observe that even if the mean training loss converges to $0$, there are a large number of atypical nodes for whom \gcn fails to adequately learn the labels. 

The limitations of GNNs are usually studied in context of their abilities to recognize isomorphic graphs or distinguish non-isomorphic graphs. Several approaches \cite{zheng2023towards} for feature-, topology- and GNN architecture enhancement have been proposed to improve the expressive power of GNNs. Nevertheless, these enhancements are usually motivated by the task of graph classification. While for graph classification, one would like to map structurally similar graphs to the similar representations, this may not always be true for the case of node-classification. 

\begin{wrapfigure}{r}{0.4\textwidth}
\vspace{-7mm}
\includegraphics[width=1.0\linewidth]{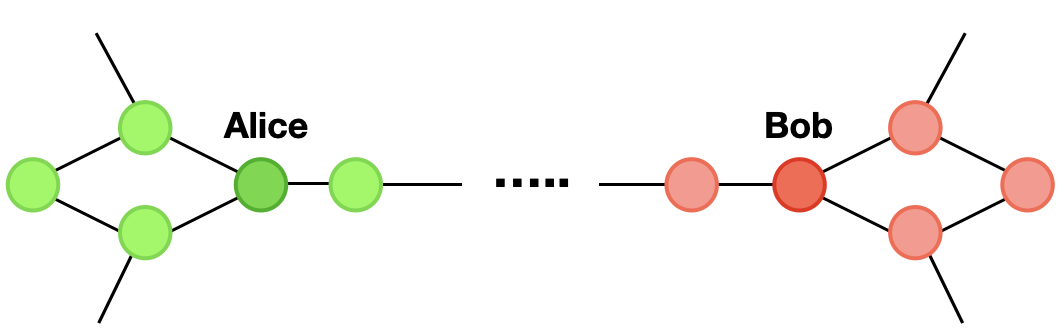} 

\caption{Example of a social network where Alice and Bob have similar local neighborhoods but different interests (labels).}
\label{fig:SN}
\vspace{-10mm}
\end{wrapfigure}
Consider for example a social network as shown in Figure \ref{fig:SN}, for which only the topological structure and the node labels are known. Two far away nodes (Alice and Bob) in such a graph may have similar local or even higher order topological structure (making their computational graphs as utilized by GNNs isomorphic) but can have completely different interests (labels).

From the above discussion we make two key observations. \textit{First}, the neighborhood aggregation approach of GNNs even of the ones which exploit higher order neighborhood is insufficient to encode node proximity information. While such an observation is not new, it has only been seen as a limitation for the task of link prediction \cite{srinivasan2019equivalence} and has been overlooked for node classification. In fact the task of link prediction and node classification are closely related for certain network types. For example, in a social network new friendships (links) may lead to adoption of new user interests (labels). In other words links could be the \textit{causal} factors for node labels. Several complex GNN architectures have been proposed to jointly encode positional and structural representations of the nodes. Nonetheless, most of the improvements are exhibited for the graph classification task and the nuances of node classification especially for the multi-label case are simply ignored.
\textit{Second}, GNNs are not able to exploit the label distribution of neighborhood nodes specifically in case of multi-label datasets. For instance, we show that a simple baseline which predicts a node labels based on a simple aggregation of known labels of its neighbors (in case they are in the training set) outperforms existing GNNs by a large margin.

\mpara{Our Contributions.} 
First, we conduct an empirical analysis of the training dynamics of existing GNN models on real world datasets, demonstrating their limited learning capabilities when applied to multi-label classification tasks.
Second, we introduce a simple yet novel framework, \mymodel, designed to fully leverage the available input information for each node—namely its features, labels, and position in the graph. Through a combination of theoretical analysis and large-scale experiments, we demonstrate that \mymodel outperforms even highly expressive GNNs, which are designed to go beyond the limitations of the 1-Weisfeiler-Lehman (1-WL) test.

\section{Background and Related Work}

\label{sec:notations}

\mpara{Notations.} Let $\mathcal{G} = (\mathcal{V}, \mathcal{E})$ denote a graph where $\mathcal{V}=\left\{v_{1}, \cdots, v_{n}\right\}$ is the set of vertices, $\mathcal{E}$ represents the set of links/edges among the vertices. We further denote the adjacency matrix of the graph by $\mathbf{A} \in\{0,1\}^{n \times n}$ and $a_{i,j}$ denotes whether there is an edge between $v_{i}$ and $v_{j}$. $\mathcal{N}(v)$ represents the immediate neighbors of node $v$ in the graph.
Furthermore, let $\mathbf{X}=\left[\mathbf{x}_{1}, \ldots, \mathbf{x}_{n}\right] \in \mathbb{R}^{n \times D}$ and $\mathbf{Y}=[\mathbf{y}_{1}, \ldots, \mathbf{y}_{n}] \in \{0,1\}^{n \times C}$ represent the feature and label matrices corresponding to the nodes in $\mathcal{V}$. In the feature matrix and label matrix, the  $i$-th row represents the feature/label vector of node $i$. Let $\ell(i)$ denote the set of labels that are assigned to node $i$. 
Finally, let $\mathcal{F}$ correspond to the feature set and $\mathcal{L}$ be the set of all labels.

\mpara{Problem Setting.} In this work, we focus on the general setting of the node classification task on graph-structured data. In particular, we are given a set of labeled nodes $\mathcal{V}_l$ and unlabelled nodes $\mathcal{V}_u$ such that each node can have arbitrary number of labels. We are then interested in predicting labels of $\mathcal{V}_u$. We assume that the training nodes are completely labeled. We deal with the transductive setting multi-label node classification problem, where the features to all nodes $\mathbf{X}$, labels of the training nodes $\mathbf{Y}_{\mathcal{V}_l}$, and the graph structure $\mathcal{G}$ are present during training.

\mpara{Graph Neural Networks and Node Classification.} In the node classification task, classical GNNs \citep{DBLP:journals/corr/HamiltonYL17, kipf2016semi, velickovic2018graph, DBLP:journals/corr/abs-1907-04931} usually take node features $\mathbf{X}$ and the adjacency matrix $\mathbf{A}$ as input. They compute node representation by recursive aggregation and transformation of feature representations of its neighbors which are then passed to a classification module
subsequently generating the probability distribution matrix $\hat{\mathbf{Y}} = [\hat{\mathbf{y}}_1,\ldots,\hat{\mathbf{y}}_n]\in \mathbb{R}^{n \times C}$ across all labels for the input nodes: $\hat{\mathbf{Y}}=f_\mathcal{\theta}(\mathbf{X}, \mathbf{A})$,
where $\mathbf{\theta}$ represents the set of parameters in the learned model. During the deployment of classical GNNs, training labels $\mathbf{Y}_{\mathcal{V}_l}$ are solely utilized for supervision in calculating the task-specific loss $\mathcal{L}$. In the node classification task, $\mathcal{L}$ is typically the cross-entropy loss defined as: $
\mathcal{L}= -\sum_{i \in \mathcal{V}_l} \mathbf{y}_i \log \left(\hat{\mathbf{y}_i}\right)
$. For the multi-label node classification, a sigmoid layer is employed as the last layer to predict the class probabilities:  $\mathbf{y}_i\leftarrow (\operatorname{sigmoid}(\mathbf{z}_{i}\mathbf{W}))$,
where $\mathbf{z}_{i}$ and $\mathbf{W}$ correspond to the node representation at the last layer and learnable weight matrix in the classification module respectively. 

\begin{definition}[Computational graph] In what follows we refer to the neighborhood graph of a node used by the GNN to obtain its feature representation as its computational graph.
\end{definition}

\begin{definition}[Label Homophily] 
\label{def:homophily}Given a multi-label graph $\mathcal{G}$, we define the label homophily $h$ of $\mathcal{G}$ (following \cite{zhao2024multilabel}) as the average of the Jaccard similarity of the label set of all connected nodes in the graph:
$r_{homo} = {1\over  |\mathcal{E}|}\sum_{(i,j)\in \mathcal{E}} {{|\ell(i) \cap \ell(j)|\over  |\ell(i) \cup \ell(j)|} }.$
\end{definition}

\subsection{Related Work}
\mpara{Label Informed GNNs.} Building upon the success of classical GNNs, recent efforts have emerged to exploit the input label information beyond just using in the training objective \citep{6751162, DBLP:journals/corr/abs-2002-06755, ZHOU2021115063}. \gcnlpa \citep{DBLP:journals/corr/abs-2002-06755} uses label propagation to learn weights to the edges and generate weighted aggregation of local features for each ego node. To exploit label co-occurence information while training node representations, \lanc \citep{ZHOU2021115063} explicitly trains label embeddings in the graph and merge the label embeddings into the aggregated feature embeddings for each node. \cite{sato2024trainingfree} augments input features with padded label vectors and proposed a training-free GNN for the node classification task in the transductive learning setting. \citet{yang2021extract} proposed an interpretable knowledge distillation method for extracting knowledge from trained GNNs into a hybrid model consisting of a GNN and a parameterized label propagation module. 

Another line of works \citep{ chen2019multi, Huang_Zhou_2021, lanchantin2019neural, Galaxc,DBLP:journals/corr/abs-1912-11757, DBLP:journals/corr/ZhuKZ17} focusing on using GNNs to exploit the relations between the data and labels in the context of multi-label learning with independent euclidean data, such as text and images are out of the scope of current work.

\mpara{Expressive Power of GNNs.}
\label{subsec:expre_gnn}
With our new insights into the empirical investigation of learning ability of GNNs for multi-label datasets, our work can be seen connected to the the theoretical efforts on quantifying the expressive power of GNNs. Usually this expressive power is assessed in terms of GNNs recognising isomorphic substructures and distinguishing between non-isomorphic substructures \citep{barcelo:hal-03356968, dwivedi2022graph, DBLP:journals/corr/abs-1910-00452, GIN, 9759979, DBLP:journals/corr/abs-2101-10320, DBLP:journals/corr/abs-2006-04330}. Such works have predominantly focused on the task of graph classification, treating each graph as an independent instance and thus bypassing the issues of inter-sample dependencies which exist for node level tasks. We address this gap by theoretically and empirically analyzing the expressive power of classical GNNs and our model design. By incorporating label and positional representation, we argue that our model, despite its simplicity, is strictly more expressive than the employed base GNN for learning feature representation.


\mpara{Data hardness analysis.}
\label{subsec:gnn_ana}
A few works \citep{DBLP:journals/corr/abs-2008-11600, siddiqui2022metadata, DBLP:journals/corr/abs-2009-10795} have also focused on characterizing data hardness for machine learning models. For example, \citep{DBLP:journals/corr/abs-2008-11600} proposed a instance-level difficulty evaluation metric and rank the difficulty of each input data for analysis. \citep{siddiqui2022metadata} curate subsets from the input datasets and monitoring the training dynamics on the them. However, to our best knowledge, none of these works have focus on the graph-structured data and GNNs. 
\section{Empirical analysis of the expressive power of GNNs}
\label{sec:ana_moti}

In order to support our arguments regarding the limited distinguishing power of existing methods for multi-labeled nodes, we analyze the training dynamics of \gcn, \idgnn, and \hgcn—three distinct approaches, each optimized for different types of graph data. \gcn performs well on high-homophily, multi-class datasets, where connected nodes tend to belong to the same class. In contrast, \hgcn is designed to be robust to the changes of levels of homophily with the most significant improvement shown in heterophilic multi-class graphs, where neighboring nodes often differ in class labels, addressing the limitations of classical GNNs. Finally, \idgnn incorporates positional information by injecting identity-aware information into the input features. 

We chose two datasets with contrasting homophily levels: \blog (low homophily) and \dblp (high homophily). We visualized the loss dynamics during training on the training data at the saved checkpoints for the selected models. The $x$-axis denotes the epoch indices where checkpoints were saved, while the $y$-axis represents the distribution of losses for each training node. We sampled $30$ uniformly spaced checkpoints throughout the training process.
We make the following observations about the expressive power of various methods.

\begin{figure}
    \centering
    \includegraphics[width=0.3\linewidth]{figs/boxplot_GCN_blogcatalog.png}
    \hfill
    \includegraphics[width=0.3\linewidth]{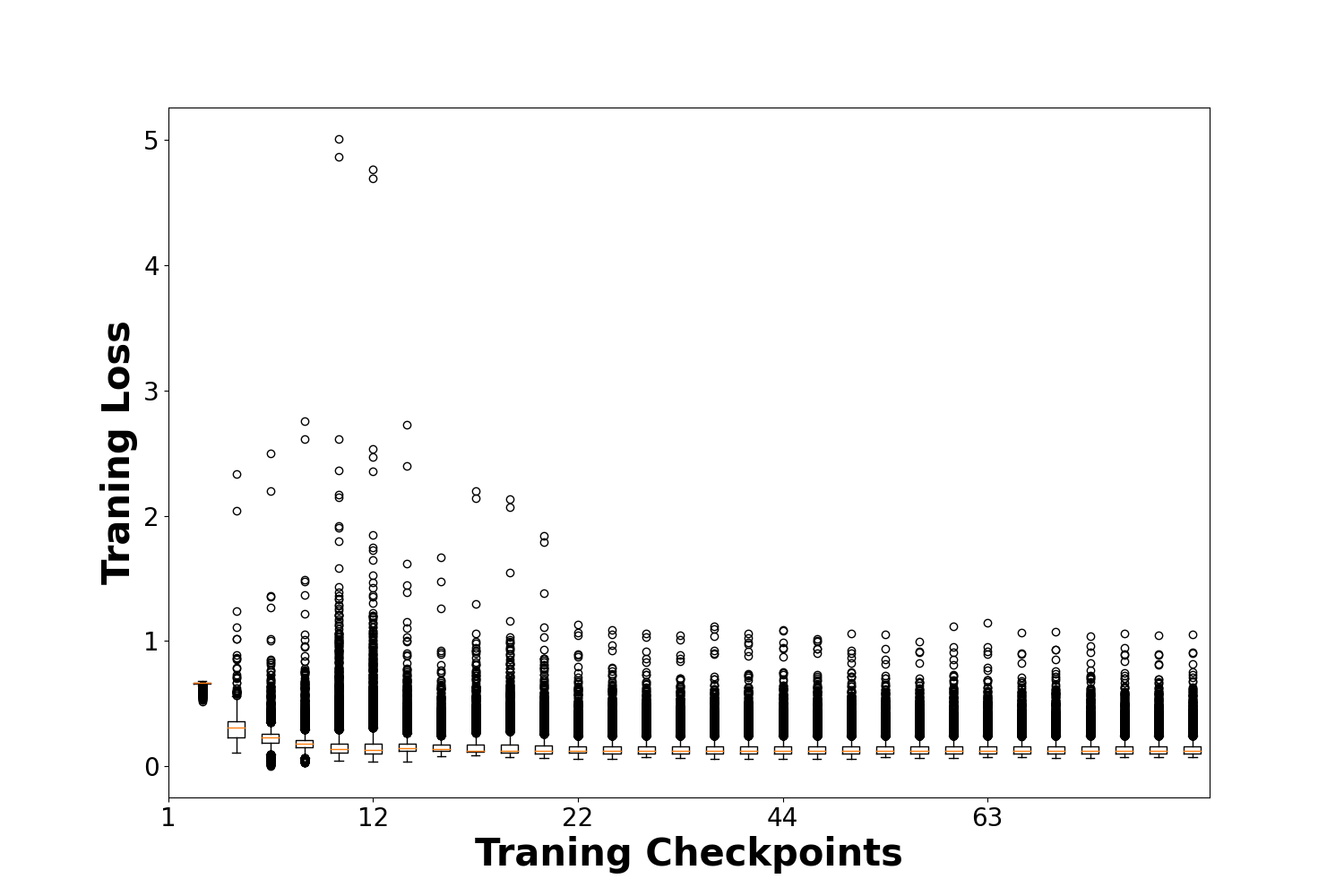}
    \hfill
    \includegraphics[width=0.3\linewidth]{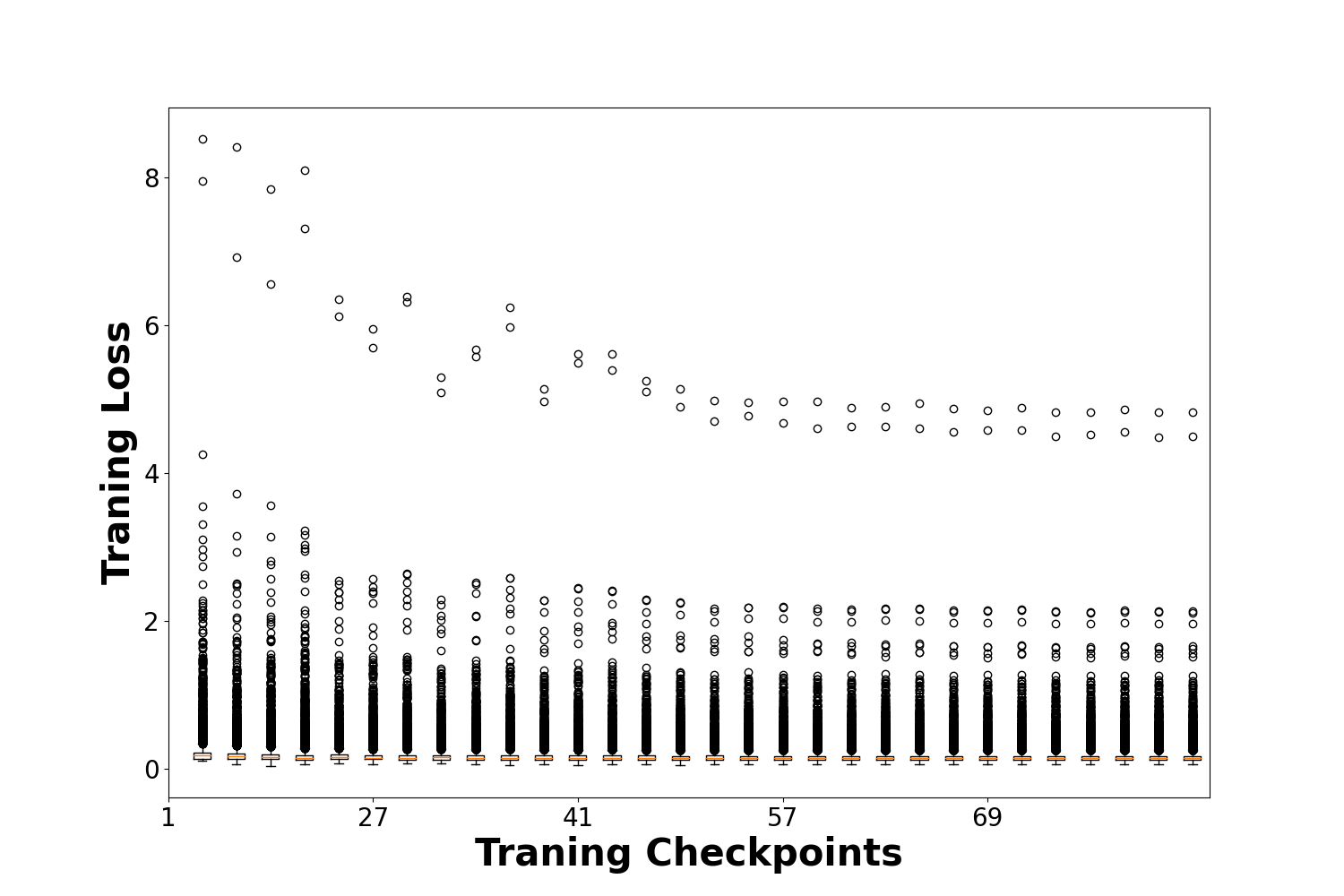}
    \caption{Visualization of training dynamics of \gcn, \hgcn, and \idgnn on the dataset \blog.}
    \label{fig:train_dym_blogcat}
\end{figure}
 \paragraph{\idgnn is not necessarily more expressive than \gcn} We consider the extreme case of \blog where no node features are available. Further we label all nodes in \blog and use them for training \gcn and \idgnn. In Figure \ref{fig:train_dym_blogcat} we both \gcn and \idgnn underfits the training data where the training loss does not converge for a large number of nodes. As in absence of node features node representations are learnt based on local neighborhood structures, it is to be expected that two nodes with similar neighborhood structures will be mapped to similar representations irrespective of their positions in the network. In such a case the expressive power of such models cannot be arbitrarily trivially by increasing the number of learnable parameters. The situation does not improve when using \idgnn which supposedly encodes additionally the positional information of the nodes. 

 \paragraph{Unexpected behavior on high homophilic datasets}

\begin{figure}
     \includegraphics[width=0.3\linewidth]{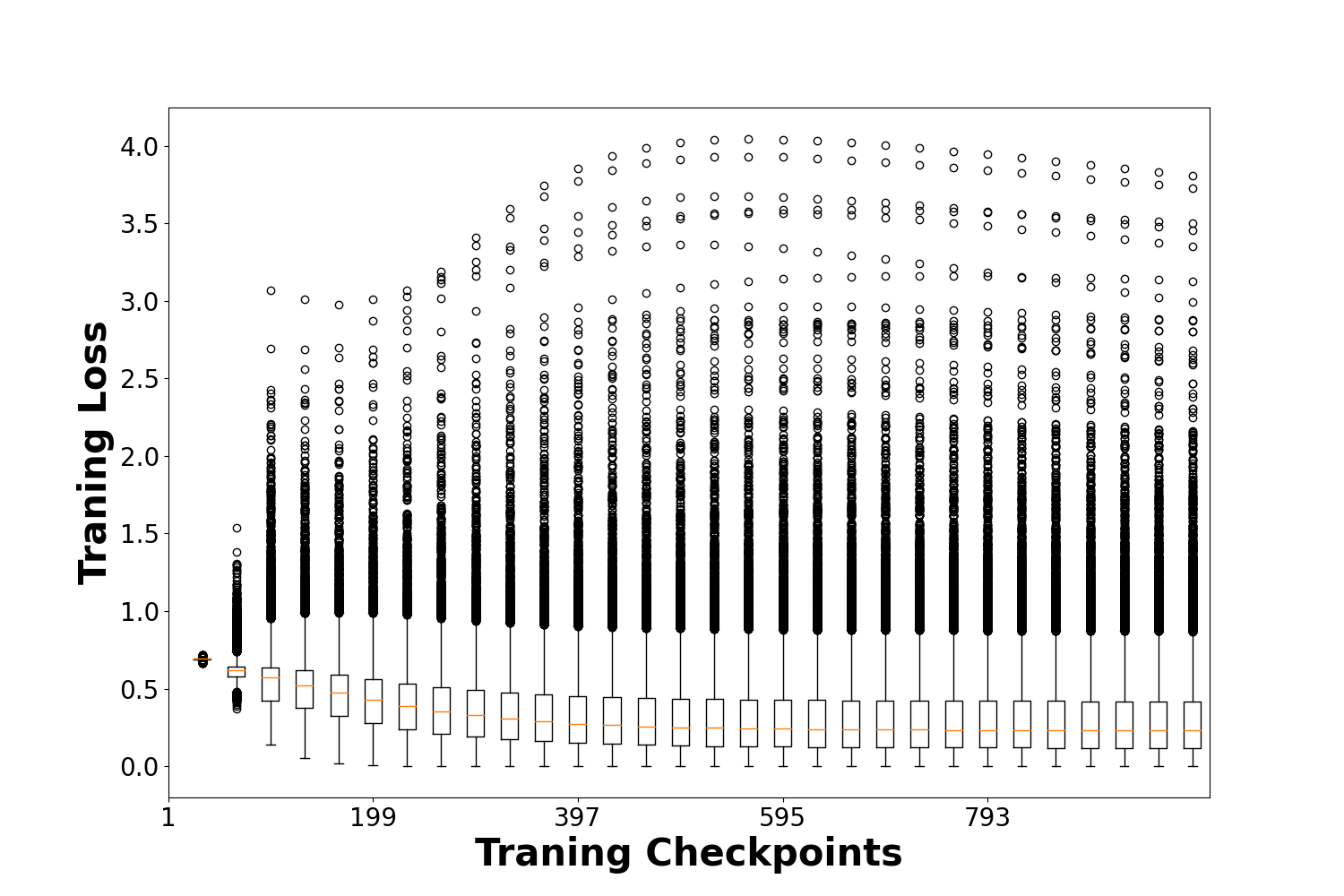}
     \hfill
     \includegraphics[width=0.3\linewidth]{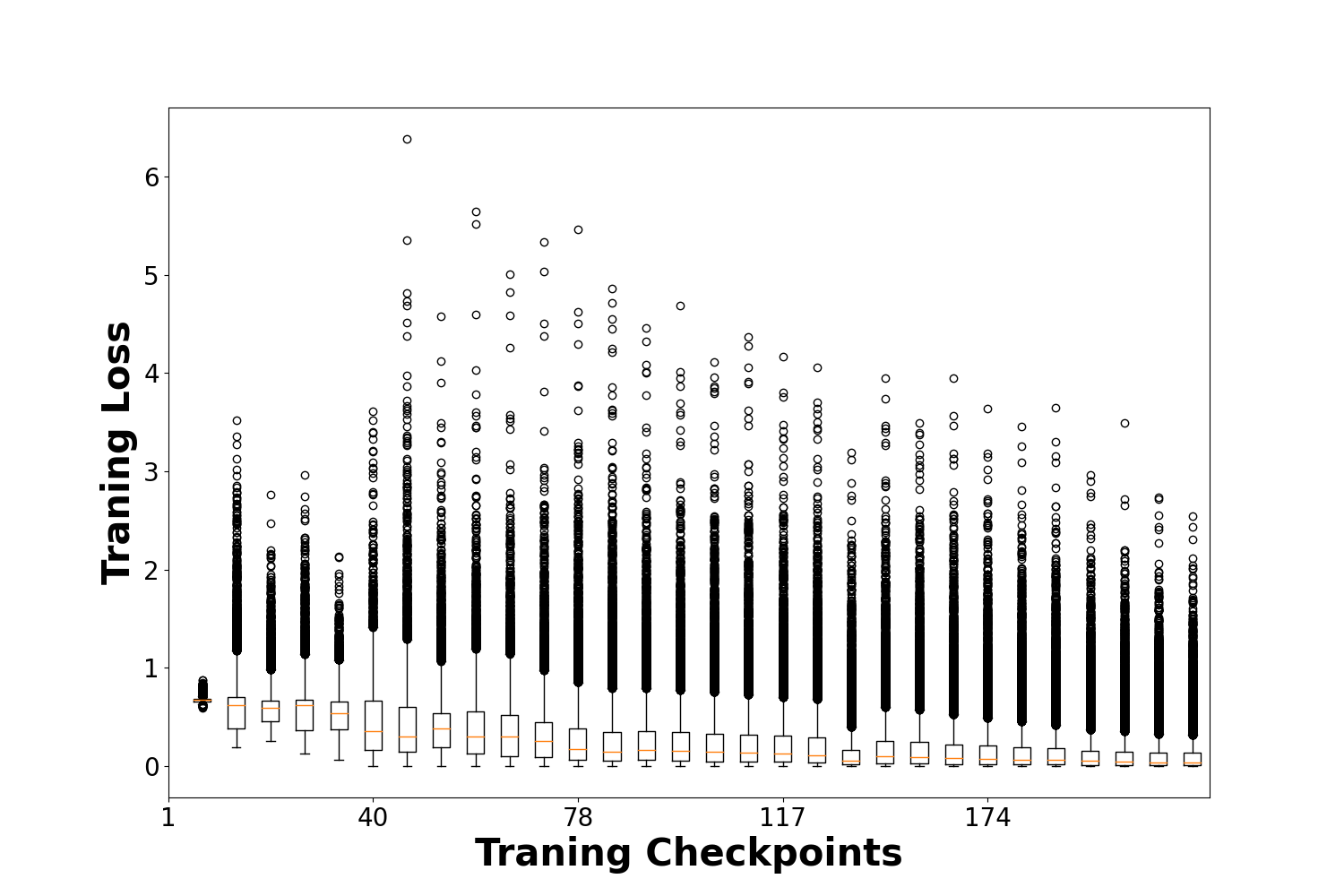}
     \hfill
     \includegraphics[width=0.3\linewidth]{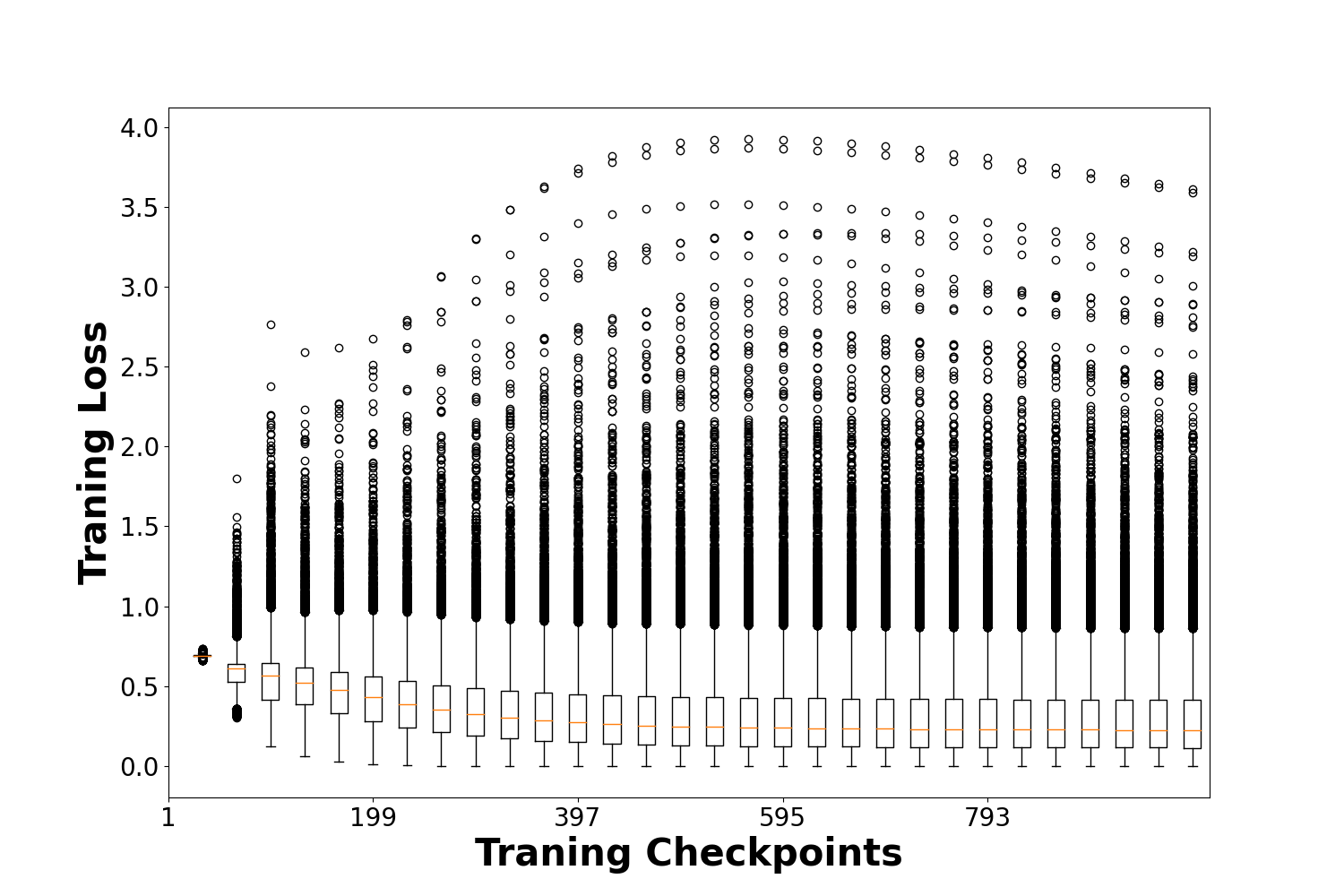}
     
     \caption{Visualization of training dynamics of \gcn, \hgcn, and \idgnn on the dataset \dblp.}
     \label{fig:train_dym_dblp}
 \end{figure}

In Figure \ref{fig:train_dym_dblp}, we compare the training dynamics of \gcn, \hgcn, and \idgnn when trained on $60\%$ of the labeled nodes of \dblp. For all three models, we observe that although the training losses decrease and converge more effectively than on \blog, due to the high label homophily, there still remains a significant number of nodes with much higher losses than the average. This suggests that the models are not learning well from these nodes. Notably, the training losses for the atypical nodes increase as the average training loss decreases, indicating that the models tend to focus on easier nodes while neglecting the harder ones.

\section{Our Approach}
\label{sec:proposed_method}

\begin{figure}

\includegraphics[width=1.0\linewidth]{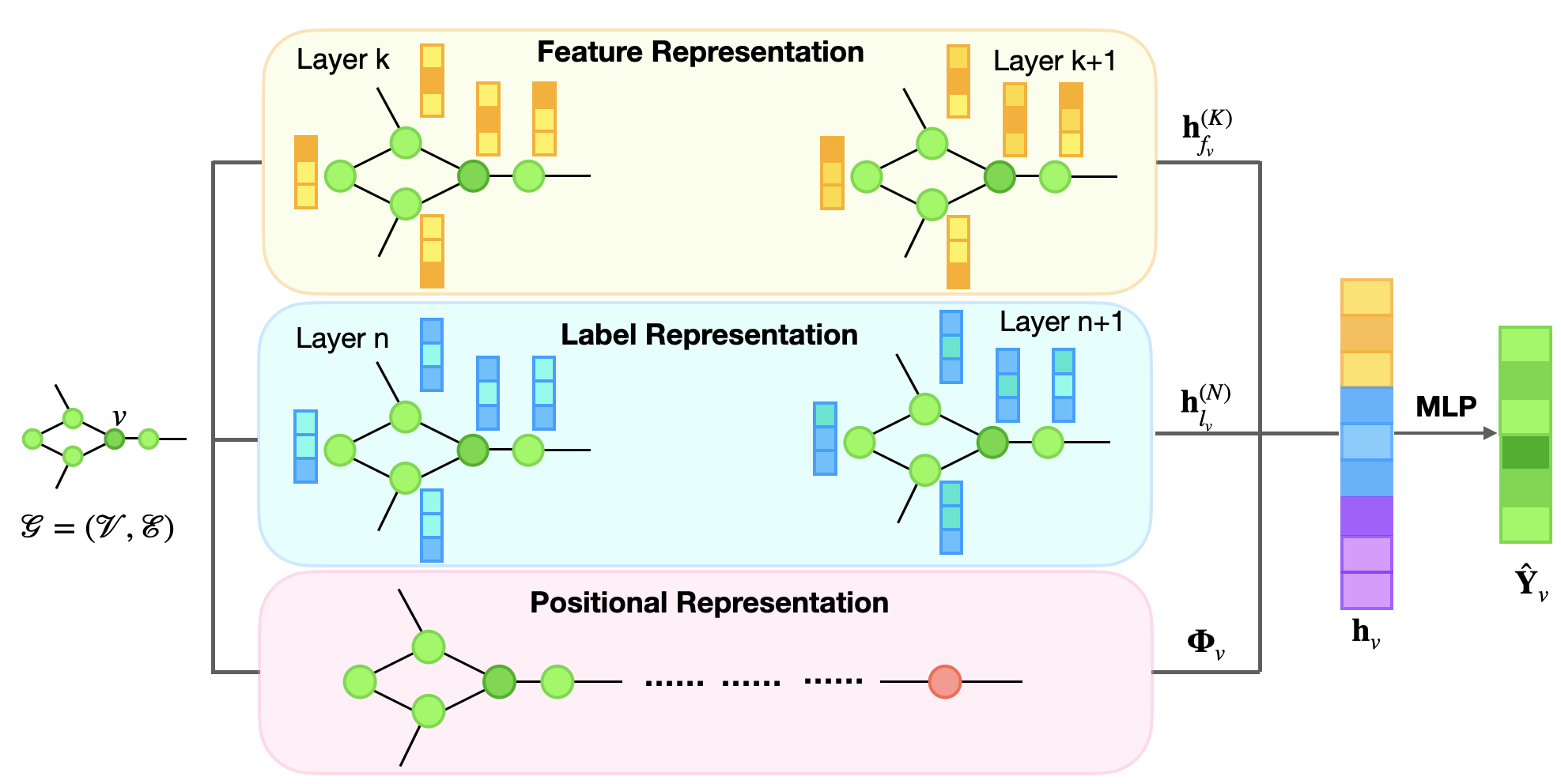}
    \caption{Illustration of the structure of \mymodel.}
    \label{fig:model_str} 
\end{figure}
As discussed in the above sections, even the most expressive GNNs in terms of passing the various graph isomorphism tests may not be able to distinguish between two  nodes with isomorphic computational graphs and  different labels. Usually the labels of such nodes are determined by the labels of their close neighborhoods rather than the overall structure of the neighborhoods they are a part of. For example in social networks two far away nodes (users) irrespective of the local structures they area part of may not have similar labels (interests). In other words in addition to the node features, the label information of the node's neihborhood as well as its relative position in the graph are the important factors influencing its labels. Our proposed methods is based on effective exploitation of such information which we detail below.



\subsection{The Proposed Method}

We design our model with three sub-modules and propose a simple yet effective method that maximizes the utility of the information in the graph, even when parts of the input are of poor quality. Additionally, we theoretically prove that the expressive power of our proposed method is strictly greater than that of the employed base GNN. In particular, our model, which we refer to as \mymodel, consists of three input processing modules, namely (i) feature representation module, (ii) label representation module (iii) node position/proximity representation module. Each of these modules are used to capture different and independent aspects of a node representation. We illustrate the overall module structure in the following Figure \ref{fig:model_str}, summarize the algorithm in Appendix \ref{app:algo} and explain the design of each modules in the following:


\subsubsection{The Feature Representation Module}

The feature representation module takes as input the node feature matrix $\mathbf{X}$ and the adjacency matrix $\mathbf{A}$ and iterate for $K$ layers and outputs node feature representation $\mathbf{h}^{(K)}_{f_v}$ for each node $v$ as:

\begin{equation}
\label{eq:fp}
\mathbf{h}^{(K)}_{f_v\_{\mathcal{N}(v)}} =\operatorname{AGG}_{(f)}^{(K)}\left(\left\{\mathbf{h}_{f_u}^{(K-1)}: u \in \mathcal{N}(v)\right\}\right), \quad
\mathbf{h}^{(K)}_{f_v}=\operatorname{COMB}_{(f)}^{(K)}\left(\mathbf{h}^{(K)}_{f_v\_{\mathcal{N}(v)}}, \, \mathbf{h}^{(K-1)}_{f_v}\right)
\end{equation}
where $\mathbf{h}^{(0)}_{f_v}$ is initialized as $\mathbf{h}^{(0)}_{f_v}= \mathbf{x}_v$ and $\operatorname{AGG}$and $\operatorname{COMB}$ are general purpose aggregation and combination functions defined over one-hop node neighborhood.
One can in general instantiate these functions corresponding to any of the existing message passing GNNs. 

In this work we obtain  representation of a node at layer k as $$\mathbf{h}^{(K)}_{f_v} = ReLU\left(\sum_{u \in \mathcal{N}(v) \cup v} \frac{1}{\sqrt{d_u d_v}} \mathbf{h}^{(K-1)}_{f_u}\right).$$



where $d$ indicate the degree of a node in the graph. Inspired by the observation \cite{zhao2024multilabel} that in multi-label graphs, despite low homophily, similar nodes are often located in the local neighborhoods we limit the feature aggregation in each layer to one-hop neighborhood.

\subsubsection{The Label Representation Module}

Secondly, we design a label representation module to generate label informative embeddings for each node by using part of the true labels of the training nodes as input. We initialize the input label representation of a node as follows 
Specifically, we initialize $\mathbf{h}^{(0)}_{l_v}$ as follows:


\begin{equation}
  \mathbf{h}^{(0)}_{l_v} = 
  \begin{cases} 
      \mathbf{y}_v, & \text{if } v \in \mathcal{V}_{\text{l}}, \\
      \mathbf{p}, & \text{if } v \in \mathcal{V}_u
  \end{cases}
\end{equation}

where $\mathbf{p} \in \mathbb{R}^{1 \times C}$ denotes uniform or zero padding. At layer $N$, the label representation of a node $v$, denoted as $\mathbf{h}^{(N)}_{l_v}$, is calculated as

\begin{equation}
\label{eq:lp}
\mathbf{h}^{(N)}_{l_v\_{\mathcal{N}(v)}} = \operatorname{AGG}_{(l)}^{(N)}\left(\left\{\mathbf{h}_{l_u}^{(N-1)}: u \in \mathcal{N}(v) \right\}\right), \quad
\mathbf{h}^{(N)}_{l_v} = \operatorname{\sigma}_{(l)}^{(N)}\left(\mathbf{h}^{(N)}_{l_v\_{\mathcal{N}(v)}}\right)
\end{equation}

where $\sigma$ represents a transformation function, such as an identity function or a learnable MLP. And the $\operatorname{AGG}$ is defined as $\mathbf{h}^{(N)}_{l_v} = \sum_{u \in \mathcal{N}(v) \cup u} \frac{1}{\sqrt{d_u d_v}} \mathbf{h}^{(N-1)}_{l_u}$. The novelty of our approach lies in integrating the label propagation algorithm (LPA) directly into the end-to-end training process for semi-supervised learning, unlike previous attempts to unify LPA with GNNs\citep{DBLP:journals/corr/abs-2002-06755}, which primarily adjusted edge weights while still relying on feature aggregation for node embeddings. 

One may argue that using true labels as inputs for the training node set, $\mathcal{V}_l$, during embedding generation generation could cause the model to assign excessive weights to self-loops, leading to overfitting and minimal learning beyond the provided labels. We address this by removing the reset step from the original label propagation algorithm, where the label vectors of the training nodes are typically reset to their true labels after each propagation step. Instead, in our framework, the label representations of the training nodes are dynamically updated by integrating information from their local neighborhoods. This forces the model to identify the most informative features for inferring the ego nodes' labels rather than merely reinforcing the initial true labels. 

Our label representation module is inspired by the observation that, in multi-label graphs, neighboring nodes often share only a subset of their labels, resulting in low homophily overall. By allowing label propagation to influence the training nodes themselves, our approach effectively captures both the commonly shared labels and the correlations within local neighborhoods, enhancing the model's ability to generalize across various nodes and label sets.

Like the feature representation module, different $\operatorname{AGG}$ and $\operatorname{COMB}$ functions can be employed for the feature and label representation modules. However, for simplicity, we use the same $\operatorname{AGG}$ and $\operatorname{COMB}$ functions as in the feature representation module. Additionally, we allow varying depths for feature propagation and label propagation within our model to capture the distinct influences of neighboring labels and features on the final representation.



\subsubsection{The Positional Encoding Module}

Thirdly, the positional representation module is designed to encode the positional information of nodes, enabling the model to differentiate between nodes with isomorphic computational graphs but different labels, even beyond their local neighborhoods. Unlike prior methods that incorporate positional encodings as additional dimensions in the input features and propagate them through local neighborhoods \citep{dwivedi2022graph}, we argue that positional encodings for multi-label classification should capture the pairwise influences between nodes. Simply aggregating this information across neighborhoods may blur critical distinctions, leading to the loss of essential positional data. Our positional encoding module is developed on two key intuitive assumptions: 

\begin{itemize}[leftmargin=*]
    \item \textbf{Distance Limitation on Influence:} 
Nodes that are far apart in the graph are less likely to influence each other, and as a result, they may belong to entirely different label sets. Our preliminary analysis of several datasets supports this assumption, indicating that it is sufficient to reconstruct the label set of a node by leveraging information from the label sets within its $k$-hop neighborhood for small $k$. 
    \item \textbf{Influence-Driven Label Similarity:} The label similarity between two nodes $u$ and $v$ increases as the potential paths for influence propagation between them increase. In other words, the  more are the opportunities for information or influence to flow between two nodes, the higher the likelihood they will share similar labels. 
\end{itemize}

To capture complementary patterns to what is already exploited by label representation module we learn node representations in an unsupervised manner without the actual information of the labels but under the above two assumptions. In general for a given pair of nodes $u$ and $v$ we are interested in generating low dimensional representations or embeddings while minimizing the following loss function
\begin{equation} 
\label{loss}
\mathcal{L}(u,v) =  (1- \mathcal{I}(u,v)) \cdot \operatorname{f}(\Phi_u,\Phi_v),\end{equation}
where $\mathcal{I}(u,v)$ quantifies the symmetrical influence of $u$ and $v$ on each other, $\operatorname{f}(.,.)$ is some function monotonically increasing in $\Phi_u\cdot \Phi_v$ . Specifically for some small $k$ we define valid node pairs to be at-most $k$-hop neighbors. For any valid pairs of nodes $u$ and $v$ we then quantify the influence $\mathcal{I}(u,v)$ as the probability of their co-occurrence in an infinite random walk over the underlying graph $\mathcal{G}$. Implementation-wise we use \deepwalk for generating node representations with short truncated walks for length 10 and window size 5. But our framework is more general and the definition of functions $\mathcal{I}(.,.)$ and $\operatorname{f}(.,.)$ can be adapted based on properties of the dataset.




\subsubsection{The Read-out Layer}
Finally, we combine the representations from the three independent modules to input in a \textit{Readout Layer} to get predictions for each node: $\mathbf{h}_v = \operatorname{COMB}(\mathbf{h}^{(K)}_{f_v}, \mathbf{h}^{(N)}_{l_v}, \Phi_v)
$, where $\Phi_v$ corresponds to the embedding vector for node $v$ generated by the node position representation module. The resulting representations are then used as input to the multi-label classifier (with output sigmoid layer as described in Section \ref{sec:notations}) which can be trained in an end-to-end manner using the binary cross entropy loss for the multi-label node classification task.

\subsection{Theoretical Analysis Of \mymodel}
We note that the feature representation module of \mymodel can be implemented using any existing GNN approach which we refer to as base GNN. In such a case \mymodel will be strictly more expressive than the the base GNN in differentiating nodes with different labels.

\begin{definition}
We call a GNN  the most expressive GNN if it can map two isomorphic graphs to the same representation and distinguish between two non-isomorphic graphs.
\end{definition}
\begin{lemma}
\label{lem:nhop}
 For any node $v \in V$, there exists an instantiation of \mymodel that encodes the label information of the training nodes in its $N$-hop neighborhood.
\end{lemma}
\vspace{-2mm}
\begin{proof}
Let $\mathbf{Y}$ be the input label matrix of the nodes padded with $0$s corresponding to labels of test nodes. We can instantiate the aggregation function in Equation \ref{eq:lp} by the uniform random walk transition matrix $\mathbf{P}=\mathbf{D}^{-1}(\mathbf{A+{\mathbf{I}}})$, where $\mathbf{D}$ corresponds to the degree matrix corresponding to the adjacency matrix (with self loops) $\mathbf{A+{\mathbf{I}}}$  and the transform function $\sigma(.)$ in Equation \ref{eq:lp} by identity. Executing the aggregation function $N$ times result in the label representation of the node $v$ as $\mathbf{h_{lv}}= \sum_{w \in \mathcal{V}}\mathbf{P}^N_{vw} \mathbf{y}_w$, where $\mathbf{y}_w$ is the label vector of node $w$. Note that all entries of $\mathbf{y}_w$ will be $0$ if $w$ is a test node.  As the element $\mathbf{P}^N_{vw} \mathbf{y}_w$ is the probability that the random walk starting from node $v$ ends at node $w$ in at most $N$ steps, the final label representation of node $v$ is the weighted sum of the label distribution of the training nodes in its $N$-hop neighbors.
\end{proof}
\begin{theorem}
\mymodel can differentiate any node that the most \textit{expressive} GNN  used for feature representation module can differentiate, while being able to differentiate certain nodes that the base GNN fails to distinguish. 
\end{theorem}
\vspace{-2mm}
\begin{proof}

It is easy to see that \mymodel will be at least as expressive as the base GNN used for implementing the feature aggregation module. For the graph without node attributes, the isomorphic nodes will be mapped to the same embedding vector as their computational graphs are isomorphic. On the other hand, \mymodel, we have two cases. In the simpler case, the label distribution of the $N$-hop neighborhood of the isomorphic nodes is known and is different. In this case by Lemma \ref{lem:nhop} the isomorphic nodes can be distinguished by their label representations. In the second case when the isomorphic nodes as well as all nodes in their N-hop neighborhood are in the test set, their positional representations can be utilized to distinguish them. For instance one can learn the positional representation of nodes in $\mathcal{G}$ by directly factorizing a node similarity matrix based on graph structure such as Katz similarity as done in \cite{ou2016asymmetric}.
\end{proof}


\section{Experimental Set-up}
\label{sec:experiment}

\mpara{Datasets.} For our experiments we employ $4$ multi-label real-world graph datasets and $2$ sets of synthetic datasets proposed in \citep{zhao2024multilabel}, where each set of the synthetic dataset consists of $5$ datasets with varying feature quality and label homophily. We summarize the notations and the characteristics of the datasets in Table \ref{tab:notation_cap} and \ref{tab:dataset}, respectively. Below we  provide a brief description of the datasets.

\paragraph{Real-world datasets} We employ (i) \blog \citep{Blogcatalog}, in which nodes represent bloggers and edges their relationships, the labels denote the social groups a blogger is a part of, (ii)\yelp \citep{DBLP:journals/corr/abs-1907-04931}, in which nodes correspond to the customer reviews and edges to their friendships with node labels representing the types of businesses (iii) \dblp \citep{DBLP:journals/corr/abs-1910-09706}, in which nodes represent authors and edges the co-authorship between the authors, and the labels indicate the research areas of the authors. and (iiii) \pcg \citep{zhao2024multilabel}, in which proteins and the protein functional interactions are represented as nodes and edges. The labels of the nodes indicate the protein phenotypes of the corresponding proteins.

\paragraph{Synthetic datasets} We obtain two types of synthetic datasets from \cite{zhao2024multilabel}. Specifically we use (i) \featqua that consists of $5$ graph datasets with the same graph structure in which the ratio of task-relevant and irrelevant feature  varies in the range of $[0.0, 0.2, 0.5, 0.8, 1.0]$ and has label homophily of $0.2$, (ii) \homolevel that consists of $5$ graph datasets in which the label homophily varies in the range of $[0.2, 0.4, 0.6, 0.8, 1.0]$. For the varying homophily experiment we only used the true node features.

\begin{table}[h]

\caption{Summary of the notations used to describe the datasets in the Table \ref{tab:dataset}.}

\begin{tabular*}{\textwidth}{c|l}
\hline
\textbf{Notation} & \textbf{Description} \\
\hline
$\mathcal{V}$ & set of nodes \\
$\mathcal{E}$ & set of edges  \\
$|\mathcal{F}|$ &the node feature dimension \\
$clus$ &the clustering coefficient \\
$r_{homo}$ &the label homophily \\
$C$ &the total number of labels  \\
$\ell_{med}$ &the median value of the number of labels per node \\
$\ell_{mean}$ &the mean value of the number of labels per node   \\
$\ell_{max}$ &the max value of the number of labels of any node\\
"N.A." &the corresponding characteristic is not available \\
"vary" &the corresponding characteristic is varying\\

\bottomrule
\end{tabular*}

\label{tab:notation_cap}
\end{table}

\begin{table}[!h]
 \caption{The characteristics of the datasets used in this work. Notations are summarized in Table \ref{tab:notation_cap}.}
 \begin{tabular}{lccccccccc}
\toprule
 \textsc{Dataset}             &  $|\mathcal{V}|$  & $|\mathcal{E}|$ &  $|\mathcal{F}|$    &  $clus$  &$r_{homo}$ & $C$ &$\ell_{med}$& $\ell_{mean}$  & $\ell_{max}$\\
\midrule
 \blog &$10K$    &$333K$    &$N.A.$  &$0.46$ &$0.10$ &$39$ &$1$ &$1.40$ &$11$ \\
 \yelp &$716K$   &$7.34M$   &$300$ &$0.09$ &$0.22$ &$100$ &$6$ &$9.44$ &$97$ \\
\dblp &$28K$ &$68K$ &$300$ &$0.61$ &$0.76$ &$4$  &$1$ &$1.18$ &$4$ \\
\pcg   &$3K$   &$37K$ &$32$ &$0.34$ &$0.17$ &$15$ &$1$ &$1.93$ &$12$ \\
\featqua &$3K$ &$2.37M$ &vary &$0.53$ &$0.2$ &$20$ &$3$ &$3.23$ &$12$ \\
\homolevel &$3K$ &vary &$10$ &vary &vary &$20$ &$3$ &$3.23$ &$12$ \\
 
\bottomrule
 \end{tabular}
 \label{tab:dataset}
 \vspace{-10pt}
\end{table}

\mpara{Baselines.} We compare our model against simple baselines targeting each of its three components, models that unify some of these components, and modules specifically designed for multi-label node classification. We categorize our compared baselines into four groups:

\textbf{(i) Simple approaches}, which include Multilayer Perceptron (\mlp\citep{haykin1994neural}), which only uses node features and ignores the graph structure, \deepwalk\citep{perozzi2014deepwalk}, which only uses graph structure and ignores the node features, and \majorityvote, which takes the direct neighbors of the test nodes which are in the training set and sum the votes of their labels. The votes are then normalized by the total number of votes in the direct neighborhood to get probabilities. 

\textbf{(iii) Classical GNNs and variants} including (a) \gcn\citep{kipf2016semi}, \gat\citep{velickovic2018graph}, and \graphsage\citep{DBLP:journals/corr/HamiltonYL17}, which are known to perform well for graphs with high label homophily, and (b)\hgcn \citep{zhu2020beyond} and \fsgnn \citep{maurya2021improvinggraphneuralnetworks}, which are designed to perform well both on homophilic and heterophilic graphs by separating the ego embedding and aggregated embeddings from different orders of neighborhoods. The difference is \fsgnn learns attention scores to important features from different hops neighborhoods. (d) \gcnlpa \citep{DBLP:journals/corr/abs-2002-06755}, which combines label propagation and GCN for node classification. 

\textbf{(ii) Label informed GNNs}, which employ convolutional operations to extract representations from node's local neighborhoods and merge them with label embeddings for the final classification. We choose \lanc \citep{ZHOU2021115063} as a baseline from this category, as it was designed for multi-label node classification, and previous works \citep{song2021semi, ZHOU2021115063} have shown its superior performance. 

\textbf{(iiii) GNN with structural and positional representations}, which integrates the positional information of the node as part of the input to obtain highly expressive models, we include \lspe \citep{dwivedi2022graph} and \idgnn \citep{DBLP:journals/corr/abs-2101-10320} from this category, which augment the input features using the positional representations and injecting the identities of the nodes, respectively.


\mpara{Variations of \mymodel.} We use $3$ variations of \mymodel in this work. Because we focus on the node classification task, in order to convert the output embeddings of the nodes into predicted probabilities for each class, we train a \mlp using the true label of the training nodes as supervision. As variations of our model, we propose (i) \mymodellin, where all the $\operatorname{AGG}$ and $\operatorname{COMB}$ functions are linear.  (ii) \mymodelmlp, where one linear layer is trained as the classifier. (iii) as \mlp can augment the expression power, we use \mymodelMlp with three layer \mlp for further improvements on the node classification task.

\mpara{Evaluation.} Motivated by \citep{dong2020towards, yang2015evaluating, zhao2024multilabel}, we compute the \textbf{Average Precision (AP) score} across three random splits of the collected datasets, with the exception of  \pcg, for which we use the predefined split as in \citep{zhao2024multilabel}.  In random splits, we allocate $60\%$ of nodes for training, $20\%$ for validation, and $20\%$ for testing purposes.\\

\section{Results and Discussion}
\label{sec:results_discussion}
In this section, we summarize the large-scale experiment results conducted on the $4$ real-world datasets as well as $2$ sets of synthetic datasets on the $12$ baselines discussed in section \ref{sec:experiment} and $3$ variations of \mymodel. Due to the space limit, we provide the hyperparameter settings and complete results with standard deviation in appendix \ref{sec:complete_results}. 

\subsection{Results On Real-world Datasets}
In Table \ref{tab:real_exres_ap}, we summarize the results on four real-world datasets. \textbf{Firstly,} it is surprising that the simple \majorityvote baseline outperforms most of the baselines methods for \blog. It further supports our hypothesis that existing GNN methods without appropriate feature information fail to effectively distinguish nodes with differing labels. Even when incorporating additional label information into the input, models like \gcnlpa fail to demonstrate improvement. Methods augmenting input features with positional encoding, such as \lspe and \idgnn, fail to surpass the performance of \deepwalk.  Notably, the simplest version of \mymodellin yields improvements of $18.4\%$ compared to the best baseline model, which underscores the effectiveness of directly integrating label information into the input and employing positional representation to distinguish nodes with differing labels. 

\textbf{Secondly}, in datasets with a high number of labels per node, such as \yelp, low homophily arises from diverse label assignments, however, nodes that share the same subset of their labels with the ego node are in the local neighborhoods. Consequently, straightforward baselines like \majorityvote outperform more complex methods such as \mlp, \deepwalk, and \gcnlpa. Interestingly, GNNs combined with sampling techniques in the local neighborhood, such as \graphsage, demonstrate superior performance compared to other designs of GNN. Nevertheless, our model (which uses GCN as base feature representation module) performs comparably well with the winning model in these scenarios. Performance of \mymodel can be further enhanced by instantiating feature representation module by \graphsage's message passing operation.

\begin{table}
\caption{Mean performance scores (Average Precision) on real-world datasets. The best score is marked in bold. The second best score is marked with underline. "OOM" denotes the "Out Of Memory" error.}
\centering
\begin{tabular}{>{}l|cccc}
\hline
 Method        &\blog        & \yelp      &\dblp      &\pcg     \\ \hline
 \mlp          &$0.043$      &$0.096$     &$0.350$    &$0.148$    \\ 
\deepwalk     &$0.190$      &$0.096$     &$0.585$    &$0.229$  \\ 
\majorityvote &$0.133$ &$0.112$ &$0.869$&$0.203$\\\hline

 \lanc         &$0.050$      &OOM         &$0.836$    &$0.185$ \\ \hline
\gcn          &$0.037$      &$0.131$     &$0.893$    &$0.210$ \\ 
\gat          &$0.041$      &$0.150$     &$0.829$    &$0.168$    \\ 
\graphsage    &$0.045$      &$\textbf{0.251}$     &$0.868$    &$0.185$  \\
 \hgcn         &$0.039$      &$0.226$     &$0.858$    &$0.192$  \\
 \fsgnn &$0.075$ &$0.093$ &$0.858$ &$0.163$\\ \hline 
 \gcnlpa       &$0.043$      &$0.116$     &$0.801$    &$0.167$ \\ \hline
 \lspe &$0.165$ &OOM &$0.590$&$\textbf{0.528}$ \\
\idgnn &$0.037$ &$0.134$ &$0.857$ &$0.235$ \\\hline
 {\mymodellin}  &$\textbf{0.225}$      &$0.200$  &$\underline{0.935}$ &$0.242$  \\
\mymodelmlp   &$\underline{0.220}$      &$0.217$           &$0.931$ &$\underline{0.254}$ \\
\mymodelMlp   &$0.127$      &$\underline{0.238}$  &$\textbf{0.942}$ &$0.252$ \\
\bottomrule
\end{tabular}
\label{tab:real_exres_ap}
\end{table}

\textbf{Thirdly} for \dblp which show very high label homophily, various GNN designs perform comparably well, surpassing simple baselines like \mlp and \deepwalk. By incorporating additional positional representation and label representation, our model outperform all the other baselines.

\textbf{Finally,} for \pcg which has low label homophily and high clustering coefficient, baselines using random walk to explore the diverse local neighborhood such as \lspe and \deepwalk outperform \mlp, and more sophisticated CNN- and GNN-based methods. The competitive performance of \majorityvote despite \pcg's low label homophily indicates that the nodes which share the similar labels with the ego node lies in the local neighborhood, which further explains the low performances of \hgcn and \fsgnn which rely on leveraging higher order neighborhood for low-homophily multi-class datasets.

\subsection{Results On Synthetic Datasets}
We also conducted the node classification task on two sets of synthetic datasets with varying feature quality and label homophily and the results are summarized in Table \ref{tab:syn_exres}. Our \mymodel outperforms all compared models baselines for most of the cases.

\begin{table}[!h]
\setlength{\tabcolsep}{2pt}
\caption{Average Precision (mean) on the synthetic datasets with varying levels of feature quality and label homophily (see definition \ref{def:homophily}). $r_{ori\_feat}$ and $r_{homo}$ refer to the fraction of original features and the varying label homophily of the graph with all true features, respectively. It was not possible to run \lspe on our synthetic datasets due to its large runtime.}
\centering
\small
\begin{tabular}{l|ccccc|ccccc}
\hline
\multirow{2}{*}{Method}        & \multicolumn{5}{c|}{$r_{ori\_feat}$} &\multicolumn{5}{c}{$r_{homo}$}\\
      & $0.0$          &$0.2$          &$0.5$            &$0.8$           &$1.0$   & $0.2$               &$0.4$                &$0.6$                &$0.8$               &$1.0$\\
\hline 
\rule{0pt}{2.5ex}\mlp       &$0.171$ &$0.185$ &$0.222$ &$0.266$ &$\mathbf{0.336}$ &$\mathbf{0.362}$ &$0.354$ &$0.355$ &$0.360$ &$0.355$\\ 
\deepwalk  &$0.181$ &$0.181$ &$0.181$ &$0.181$ &$0.181$  & $0.181$  &$\underline{0.522}$   &$\mathbf{0.813}$   &$0.869$ &$0.552$\\ 
\lanc  &$0.191$ &$0.191$ &$0.192$ &$0.211$ &$0.215$  &$0.179$ &$0.397$ &$0.473$ &$0.500$ &$0.634$\\\hline
\gcn   &$0.261$ &$\underline{0.261}$ &$\underline{0.261}$ &$0.262$ &$0.262$ &$0.261$ &$0.308$ &$0.360$ &$0.430$ &$0.450$\\ 
\gat   &$0.171$ &$0.169$ &$0.176$ &$0.177$ &$0.182$ &$0.196$ &$0.341$ &$0.353$ &$0.399$ &$0.403$ \\ 
\graphsage &$0.171$&$0.191$&$0.232$&$\underline{0.268}$&$\underline{0.304}$ &$\underline{0.317}$ &$0.408$ &$0.437$ &$0.484$ &$0.491$ \\ 
\hgcn      &$0.182$&$0.202$&$0.224$&$0.258$ &$0.290$ &$0.300$ &$0.487$ &$0.519$ &$0.617$ &$0.618$\\
\fsgnn &$0.180$  &$0.191$ &$0.211$ &$0.228$ &$0.243$ &$0.240$    &$0.433$   &$0.419$   &$0.328$   &$0.371$\\\hline
\gcnlpa    &$0.174$ &$0.175$ &$0.170$&$0.170$&$0.172$ &$0.174$ &$0.277$ &$0.352$ &$0.414$ &$0.346$\\ \hline
\idgnn &$0.261$ &$\underline{0.261}$ &$\underline{0.261}$ &$0.261$ &$0.261$ &$0.261$ &$0.308$ &$0.363$ &$0.432$ &$0.457$\\ \hline

\mymodellin &$0.249$ &$0.248$ &$0.248$ &$0.248$ &$0.247$ &$0.256$&$0.468$ &$0.687$ &$0.812$ &$0.445$\\
\mymodelmlp &$\mathbf{0.289}$ &$\mathbf{0.283}$ &$\mathbf{0.283}$ &$\mathbf{0.280}$ &$0.288$ &$0.288$ &$0.492$ &$0.701$ &$\underline{0.910}$ &$\mathbf{0.832}$\\
\mymodelMlp &$\underline{0.270}$ &$\underline{0.261}$ &$0.255$ &$0.254$ &$0.267$ &$0.269$ &$\mathbf{0.527}$ &$\underline{0.795}$ &$\mathbf{0.916}$ &$\underline{0.811}$\\
\hline
\end{tabular}
\label{tab:syn_exres}
\end{table}

\mpara{Varying Feature Quality.} Across all levels of feature quality, the performance of \deepwalk remains consistent, as it generates node representations solely based on the graph structure. The \mlp is predictably the most sensitive method to changes in feature quality, as it completely disregards the graph structure. \lanc also demonstrates sensitivity to variations in feature quality, extracting feature vectors from the local neighborhood through convolutional operations on the stacked feature matrix of direct neighbors. Among the classical GNNs and their variations, \graphsage and \hgcn exhibit the most significant improvements with increasing feature quality. With the integration of label propagation and positional encoding, \gcnlpa and \idgnn show limited but stable performance across varying feature qualities. In contrast, our model and its three variants achieve superior and robust performance across different levels of feature quality, with the most notable improvement observed at the lowest feature quality. Notably, in the synthetic datasets where the input features are strongly correlated with the labels \cite{zhao2024multilabel}, \mlp outperforms all other baselines at the highest feature quality level.

\mpara{Varying Label Homophily.} The performance of the \mlp remains relatively consistent across different levels of label homophily, primarily because it relies solely on input features. Conversely, while considerable effort has been invested in developing complex methods for node classification tasks, we find that simpler baselines like \deepwalk surpass standard GNNs in several instances. Specifically, on synthetic datasets with label homophily levels of $0.4$ and $0.6$, \deepwalk emerges as the top-performing method. Given that \lanc extracts feature vectors from the local neighborhood through convolutional operations on the stacked feature matrix of direct neighbors, it is not surprising that its performance is highly sensitive to changes in label homophily. All classical GNNs and their variants exhibit similar sensitivity to variations in label homophily, demonstrating improved performance under high label homophily conditions. Although \gcnlpa and \idgnn integrate label information to adjust edge weights and enhance input features with positional encoding, their improvements over simple classical GNNs are limited. In contrast, our model and its variants demonstrate significant enhancements. Notably, at the lowest level of homophily, where neighbors are quite dissimilar compared to ego nodes and input features are strongly correlated with the labels, the \mlp outperforms all other baselines.

\mpara{Performance of majority vote.} In Table \ref{tab:limitations}
we provide performance of \majorityvote baseline for our synthetic datasets. 
Unlike for real world datasets majority vote baseline outperforms all methods. For low homophilic synthetic graphs we attribute this finding to high edge density of the synthetic graphs when the homophily is low. Note that the edge density  is close to $0.27$ for \featqua which has homophily level of $0.2$. This turns out to be on average around $800$ 1-hop neighbors per node when the maximum number of labels per node is $12$.

While the edge density decreases with increasing homophily, the performance of \majorityvote still stands out to the best. To understand this phenomena consider the extreme case of homophily 1. On average each has $15$ neighbors but all of its neighbors have the same label set as the node itself. The problem then becomes trivial for \majorityvote baseline which can give the correct answer even if only one node in the neighborhood is labelled. Interestingly while \majorityvote is supposedly a weak baseline in real world datasets it provides an upper bound of performance on chosen synthetic datasets. Our method nevertheless outperforms other baselines in most of the settings. For instance, at the highest homophily level of $1$ for \homolevel dataset we achieve average precision (0.832) comparable to the \majorityvote  (0.847) and an improvement of $31.23\%$ over the second best method (\lanc).

\begin{table}[!h]
\setlength{\tabcolsep}{5pt}
\caption{Average Precision (mean) on the synthetic datasets with varying levels of feature quality and label homophily when \majorityvote is employed. }
\centering
\begin{tabular}{l|ccccc|ccccc}
\hline
\small
\multirow{2}{*}{Method}        & \multicolumn{5}{c|}{$r_{ori\_feat}$} &\multicolumn{5}{c}{$r_{homo\_clean}$}\\
      & $0.0$          &$0.2$          &$0.5$            &$0.8$           &$1.0$   & $0.2$               &$0.4$                &$0.6$                &$0.8$               &$1.0$\\
\hline  
\majorityvote &$0.477$ &$0.477$ &$0.477$ &$0.477$ &$0.477$ &$0.477$ &$0.926$ &$0.957$ &$0.938$ &$0.847$\\\hline

\hline
\end{tabular}
\label{tab:limitations}
\end{table}

\section{Ablation Study}

In this section, we verify the effectiveness of our design by disabling one module at a time in \mymodellin. There are three module in \mymodel, the feature representation module, the label representation module, and the positional encoding module. Thus, we generate three variations of \mymodellin, i.e., \mymodellin without feature representation module, denoted as \mymodellin w/o FR, \mymodellin without label representation module, denoted as \mymodellin w/o LR, and \mymodellin without positional encoding module, denoted as \mymodellin w/o PE and test the performances of them on the dataset \blog and \dblp and summarize the results in the table \ref{tab:ablation}.

\begin{table}
\caption{Ablation study on \mymodellin by disabling each module at a time on the dataset \blog and \dblp.}
    \begin{tabular}{c|c|c}
        \hline
         & \blog &\dblp \\ \hline
        \mymodellin w/o FR &$0.225$ &$0.920$ \\ \hline
        \mymodellin w/o LR &$0.225$ &$0.864$\\\hline
        \mymodellin w/o PE &$0.045$ &$0.934$\\\hline
        \mymodellin &$0.225$ &$0.935$ \\ \hline
    \end{tabular}
    \label{tab:ablation}
\end{table}

\blog is a dataset without input features and exhibits the lowest label homophily. As a result, disabling the feature and label representation modules has little effect on the final performance. However, disabling the positional encoding module leads to a significant performance drop. This indicates that when input features are poor, and the labels in the local neighborhood provide little useful information for inferring the labels of a node, \mymodel adapts by relying more on the positional encoding module. In contrast, \dblp, a dataset with the highest label homophily, shows that input features are limited in their ability to infer node labels, as seen in the weaker performance of \mlp on this dataset. Here, disabling the label propagation module causes the largest performance drop. By comparing these results on \blog and \dblp with differing characteristics, we demonstrate that our model effectively identifies and leverages the most informative aspects of the input to predict node labels.





\section{Conclusion}
 We delve into the often-overlooked scenario of multi-label node classification, revealing that GNNs struggle to learn from multi-label datasets, even with abundant training data. We show that even the most expressive (in terms of distinguishing between non-isomorphic graphs) GNNs may fail to effectively distinguish nodes with different label-set without node attributes and explicit label information. Our simple yet effective approach, \mymodel, which leverages feature, label, and positional information of  a node to predict its labels consistently outperforms existing methods on multi-label node classification task. 

\bibliographystyle{plainnat}
\bibliography{main}
\clearpage

\appendix
\section{Appendix}

\textbf{Organization of the Appendix. } In Section \ref{app:algo}, we provide pseudo code for the algorithm of \mymodel. In the following Section \ref{app:sup_ma}, we provide supplementary materials about the experiments conducted in this work. Specifically, in \ref{app:data}, we summarize the detailed characteristics of the datasets used in this work. Furthermore, The hyperparameter setting to reproduce the results in this work are summarized in Section \ref{sec:hyper_base} and \ref{sec:hyper_my}. We provide the complete results with standard deviation on the three random splits described in the main paper in section \ref{sec:complete_results}. Last but not least, we provide broader impact of our paper in Appendix \ref{app:broader_impact}.


\subsection{Pseudocode for \mymodel}
\label{app:algo}

In this section, we summarize the general framework for the node classification task on graph-structured data, which we refer to as \mymodel. As shown in Algorithm \ref{algo:my_model} from line $2$ to $5$ and $6$ to $9$, for each node $v \in \mathcal{V}$, we first perform $K$ layers of feature propagation and $N$ layer of label propagation to generate feature representation and label representation, respectively. Note that any aggregation and message-passing functions from classical GNNs can be used as $\operatorname{AGG}$ and $\operatorname{COMB}$, and they are not restricted to be identical for feature and label propagation, which makes our method a general framework for the node classification task, as any other GNNs can be used as backbone model. In the implementation used in the experiment Section \ref{sec:experiment}, we use the operation $\operatorname{AGG}$ and $\operatorname{COMB}$ from \gcn.

Then, we generate $N_{rm}$ number of random walks and train a $SkipGram$ model to obtain the positional embedding $\phi_v$. In this stage, the embeddings of the nodes that are far away from each other in the graph are forced to be different. 

In line $11$, we $\operatorname{COMB}$ the representations generated from input feature, input label, and the positional information to obtain the overall representation for node $v$, denoted as $\mathbf{h}_v$.

Finally, we feed $\mathbf{h}_v$ into a $\operatorname{ReadOut}$ layer, which is typically a MLP with \textit{Sigmoid}, to obtain the predicted probabilities.

\begin{algorithm}[!h]
\caption{Learning node representations with \mymodel}
\label{algo:my_model}
\begin{algorithmic}[1]
\Require{graph $\mathcal{G}=(\mathcal{V},\mathcal{E})$; input node features $\mathbf{X}$; padded label matrix $\mathbf{h}^{(0)}_{l_v}$; number of feature propagation $K$; number of label propagation $N$; Aggregation function $\operatorname{AGG(.)}$; Combine function $\operatorname{COMB(.)}$; Read-out layer $\operatorname{ReadOut(.)}$}
\Ensure{Prediction matrix $\mathbf{\hat{Y}}_v \in \mathbb{R}^{n\times C}$}
\For{$v \in \mathcal{V}$}
    \State{Initialize input features $\mathbf{h}^{(0)}_{f_v} = \mathbf{X}$}
    \For{$i \in K$}
    \State{$\mathbf{h}^{(i)}_{f_v\_{\mathcal{N}(v)}}=\operatorname{AGG}_{(f)}^{(i)}\left(\left\{\mathbf{h}_{f_u}^{(i-1)}: u \in \mathcal{N}(v)\right\}\right)$}
        \State{$\mathbf{h}^{(i)}_{f_v}=\operatorname{COMB}_{(f)}^{(i)}\left(\mathbf{h}^{(i)}_{f_v\_{\mathcal{N}(v)}}, \, \mathbf{h}^{(i-1)}_{f_v}\right)$}
    \EndFor

    \State{Initialize input label matrix $\mathbf{h}^{(0)}_{l_v}$}
    \For{$j \in N$}
        \State{$\mathbf{h}^{(j)}_{l_v\_{\mathcal{N}(v)}}=\operatorname{AGG}_{(l)}^{(j)}\left(\left\{\mathbf{h}_{l_u}^{(j-1)}: u \in \mathcal{N}(v)\right\}\right)$}
        \State{$\mathbf{h^{(j)}_{l_v}}=\operatorname{\sigma}_{(l)}^{(N)}\left(\mathbf{h}^{(j)}_{l_v\_{\mathcal{N}(v)}}\right)$}
    \EndFor
    \State{Generate positional embeddings $\Phi_v$ }
    \State{Generate input representation for node $v$: $\mathbf{h}_v = \operatorname{COMB}(\mathbf{h}^{(K)}_{f_v}, \mathbf{h}^{(N)}_{l_v}, \mathbf{\Phi}_v)$}
    \State{$\mathbf{\hat{Y}_v} = \operatorname{ReadOut}\left( \mathbf{h}_v\right)$}
    
    \EndFor 
    \State{$\mathbf{\hat{Y}}:=[ \mathbf{\hat{Y}}_{v_1}, \mathbf{\hat{Y}}_{v_2},\ldots,\mathbf{\hat{Y}}_{v_n} ]$}
    \end{algorithmic}
\end{algorithm}

\subsection{Supplementary Material For Experiments}
\label{app:sup_ma}

\subsubsection{Description of Datasets}
\label{app:data}



The detailed statistics of the datasets are provided in Table \ref{tab:dataset_complete}.
\begin{table}[h!] 
 \caption{$|\mathcal{V}|$, $|\mathcal{E}|$, $|\mathcal{F}|$ denote the number of nodes, edges and the node feature dimension respectively. $clus$ and $r_{homo}$ denote the clustering coefficient and the label homophily (average of Jaccard similarity of labels over all edges). $C$ is the total numebr of labels. 
$\ell_{med}$, $\ell_{mean}$, and $\ell_{max}$ specify the median, mean, and max values corresponding to the number of labels of a node.  `$25$\%', `$50$\%', and `$75$\%' corresponds to the $25$th, $50$th, and $75$th percentiles of the sorted list of the number of labels for a node. "N.A." and "vary" indicate the corresponding characteristic is not available or varying in the graphs.}
\setlength{\tabcolsep}{3pt}
\centering
 \begin{tabular}{lccccccccccccc}

\toprule
 \textsc{Dataset}             &  $|\mathcal{V}|$  & $|\mathcal{E}|$ &  $|\mathcal{F}|$    &  $clus$  &$r_{homo}$ & $C$ &$\ell_{med}$& $\ell_{mean}$  & $\ell_{max}$& $25$\% & $50$\% & $75$\%\\
\midrule
 \blog &$10K$    &$333K$    &$N.A.$  &$0.46$ &$0.10$ &$39$ &$1$ &$1.40$ &$11$ &$1$ &$1$ &$2$\\
 \yelp &$716K$   &$7.34M$   &$300$ &$0.09$ &$0.22$ &$100$ &$6$ &$9.44$ &$97$ &$3$ &$6$ &$11$ \\
\dblp &$28K$ &$68K$ &$300$ &$0.61$ &$0.76$ &$4$  &$1$ &$1.18$ &$4$ &$1$ &$1$ &$1$ \\
\pcg   &$3K$   &$37K$ &$32$ &$0.34$ &$0.17$ &$15$ &$1$ &$1.93$ &$12$ &$1$ &$1$ &$2$\\

\featqua &$3K$ &$2.37M$ &vary &$0.53$ &$0.2$ &$20$ &$3$ &$3.23$ &$12$ &$1$ &$3$ &$5$\\
\homolevel &$3K$ &vary &$10$ &vary &vary &$20$ &$3$ &$3.23$ &$12$ &$1$ &$3$ &$5$\\
 
 \hline
 \end{tabular}
 \vspace{1mm}
 
 \label{tab:dataset_complete}
 \vspace{-5mm}
\end{table}

\subsubsection{Hyperparameter Setting For Baselines}
\label{sec:hyper_base}
For a fair comparison, we also tuned the hyperparameters of the collected baselines and summarized the optimal settings for each dataset. Specifically, we tuned the learning rate parameters in \{$0.1, 0.01, 0.001, 0.0001$\} and the hidden units for neural networks in \{$64, 128, 256, 512$\}. Additionally, we adopted a common deployment of neural networks with $2$ layers and used early stopping with a patience of $100$ for all methods in this work, along with a weight decay of $5e^{-4}$. The specific parameters of the collected baseline models are fixed to the setting in the original paper. We release all the detailed hyperparameter setting in \url{https://anonymous.4open.science/r/Graph-MultiFix-4121}.

\subsubsection{Hyperparameter Setting For \mymodel}
\label{sec:hyper_my}

We summarize the hyperparameter setting used in the variants of \mymodel for reproducing the experimental results in this work according to each dataset in Table \ref{tab:hyper_para_my_model}. We use the consistant notations as in the main paper. $k$ represents the number of feature propagation, $N$ indicates the number of label propagation. We also used early stopping with the patience specified in the table to prevent the models from overfitting. The hidden dimension are used for both feature representation module and label representation module. And $\Phi_{dim}$ indicates the dimension of positional representations.

\begin{table}[!h]
\setlength{\tabcolsep}{3pt}
\caption{The hyperparameter setting for variations of \mymodel.}
\centering
\begin{tabular}{c|c|c|c|c|c|c|c}
\hline \hline
& &$K$ &$N$ &patience &learning rate &hidden dim &$\Phi_{dim}$\\\hline
\multirow{4}{*}{\mymodellin} 
&\blog  &$2$ &$1$ &$100$ &$0.03$ &$256$ &$64$\\
&\yelp  &$2$ &$25$ &$100$ &$0.005$ &$256$ &$64$\\
&\dblp &$2$ &$2$ &$100$ &$0.01$ &$256$ &$64$\\
&\pcg &$2$ &$1$ &$100$ &$0.01$ &$256$ &$64$\\ \hline

\multirow{4}{*}{\mymodelmlp} 
&\blog  &$2$ &$1$ &$100$ &$0.01$ &$256$ &$64$\\
&\yelp  &$2$ &$25$ &$100$ &$0.005$ &$256$ &$64$\\
&\dblp &$2$ &$2$ &$100$ &$0.01$ &$256$ &$64$\\
&\pcg &$2$ &$5$ &$100$ &$0.01$ &$256$ &$64$\\ \hline

\multirow{4}{*}{\mymodelMlp} 
&\blog  &$2$ &$1$ &$100$ &$0.01$ &$256$ &$64$\\
&\yelp &$2$ &$25$ &$100$ &$0.005$ &$256$  &$64$\\
&\dblp &$2$ &$2$ &$100$ &$0.01$ &$256$ &$64$\\
&\pcg &$2$ &$5$ &$100$ &$0.01$ &$256$ &$64$\\ \hline
\hline

\end{tabular}

\label{tab:hyper_para_my_model}
\end{table}

\subsubsection{Complete Experiment Results}
\label{sec:complete_results}

Due to the page limit, we provide the full experiment results with standard deviation of three random splits on the collected real-world datasets and the synthetic datasets in the following Table \ref{tab:real_exres_ap_full}, \ref{tab:syn_exres_feat_full} and \ref{tab:syn_exres_homo_full}, respectively.

\begin{table}[!h]
\setlength{\tabcolsep}{7pt}
\caption{Mean performance scores (Average Precision) on real-world datasets. The best score is marked in bold. The second best score is marked with underline. "OOM" denotes the "Out Of Memory" error.}
\centering
\small
\begin{tabular}{>{\footnotesize}l|cccc}
\hline
 Method        &\blog                & \yelp                    &\dblp                        &\pcg     \\ \hline
 \mlp          &$0.043 \pm 0.63$     &$0.096 \pm 0.01$          &$0.350 \pm 0.14$             &$0.148 \pm 0.60$    \\ 
\deepwalk      &$0.190 \pm 0.17$     &$0.096 \pm 0.02$          &$0.585 \pm 0.25$             &$0.229 \pm 1.00$  \\ 
\majorityvote  &$0.133 \pm 0.00$ &$0.112 \pm 0.00$     &$0.869 \pm 0.00$                   &$0.203 \pm 0.00$                                \\\hline

 \lanc         &$0.050 \pm 0.07$     &OOM                       &$0.836 \pm 0.98$             &$0.185 \pm 1.14$ \\ \hline
\gcn           &$0.037 \pm 0.04$     &$0.131 \pm 0.06$          &$0.893 \pm 0.24$             &$0.210 \pm 0.34$ \\ 
\gat           &$0.041 \pm 0.09$     &$0.150 \pm 0.07$          &$0.829 \pm 0.16$             &$0.168 \pm 2.17$    \\ 
\graphsage     &$0.045 \pm 0.12$     &$\mathbf{0.251 \pm 0.31}$ &$0.868 \pm 0.18$         &$0.185 \pm 0.29$  \\
\hgcn          &$0.039 \pm 0.05$     &$0.226 \pm 0.51$          &$0.858 \pm 0.64$             &$0.192 \pm 0.49$  \\
\fsgnn &$0.075\pm0.00$ &$0.093\pm0.00$ &$0.858\pm0.01$ &$0.163\pm0.01$\\\hline 
\gcnlpa        &$0.043 \pm 0.31$     &$0.116 \pm 0.74$          &$0.801 \pm 1.24$             &$0.167 \pm 0.14$ \\ \hline
\lspe          &$0.165 \pm 0.00$              &OOM                       &$0.590 \pm 0.00$                      &$\mathbf{0.528 \pm 0.05}$ \\
\idgnn         &$0.037 \pm 0.00$              &$0.134 \pm 0.00$                   &$0.857 \pm 0.00$                      &$0.235 \pm 0.01$ \\\hline
{\mymodellin}  &$\mathbf{0.225 \pm 0.01}$    &$0.201\pm0.01$                   &$\underline{0.935\pm0.00}$          &$0.242\pm0.01$  \\
\mymodelmlp    &$\underline{0.220 \pm 0.00}$  &$0.217\pm0.00$                   &$0.931\pm0.00$                      &$\underline{0.254\pm 0.01}$ \\
\mymodelMlp    &$0.127 \pm 0.00$              &$\underline{0.237\pm0.00}$       &$\mathbf{0.942\pm 0.01}$           &$0.252\pm0.01$ \\
\hline
\end{tabular}
\label{tab:real_exres_ap_full}
\end{table}

\begin{table}[!h]
\setlength{\tabcolsep}{3pt}
\caption{Average Precision (mean) with standard deviation on the synthetic datasets with varying levels of feature quality. $r_{ori\_feat}$ refers to the fraction of original features of the graph with all true features. It was not possible to run \lspe on our synthetic datasets due to its large runtime.}
\centering
\small
\begin{tabular}{l|ccccc}
\hline
\multirow{2}{*}{Method}        & \multicolumn{5}{c}{$r_{ori\_feat}$}\\
      & $0.0$          &$0.2$          &$0.5$            &$0.8$           &$1.0$  \\
\hline 
\rule{0pt}{2.5ex}\mlp       &$0.171\pm0.01$ &$0.185\pm0.00$ &$0.222\pm0.00$ &$0.266\pm0.01$ &$0.336\pm0.00$ \\ 
\deepwalk  &$0.181\pm0.71$ &$0.181\pm0.71$ &$0.181\pm0.71$ &$0.181\pm0.71$ &$0.181\pm0.71$  \\ 
\majorityvote &$\mathbf{0.477\pm0.02}$ &$\mathbf{0.477\pm0.02}$ &$\mathbf{0.477\pm0.02}$ &$\mathbf{0.477\pm0.02}$ &$\mathbf{0.477\pm0.02}$ \\\hline
\lanc  &$0.191\pm0.02$ &$0.191\pm0.02$ &$0.192\pm0.01$ &$0.211\pm0.02$ &$0.215\pm0.02$  \\
\gcn   &$0.261\pm0.01$ &$0.261\pm0.01$ &$0.261\pm0.01$ &$0.262\pm0.01$ &$0.262\pm0.01$ \\ 
\gat   &$0.171\pm0.00$ &$0.169\pm0.00$ &$0.176\pm0.00$ &$0.177\pm0.00$ &$0.182\pm0.01$  \\ 
\graphsage &$0.171\pm0.00$  &$0.191\pm0.00$    &$0.232\pm0.01$    &$0.268\pm0.00$    &$\underline{0.304\pm0.01}$ \\ 
\hgcn      &$0.182\pm0.00$ &$0.202\pm0.01$    &$0.224\pm0.00$    &$0.258\pm0.01$    &$0.290\pm0.01$ \\
\fsgnn &$0.180\pm0.00$  &$0.191\pm0.00$ &$0.211\pm0.00$ &$0.228\pm0.01$ &$0.243\pm0.00$\\
\gcnlpa    &$0.174\pm0.00$     &$0.175\pm0.00$    &$0.170\pm0.00$    &$0.170\pm0.00$    &$0.172\pm0.01$ \\ 
\idgnn &$0.261\pm0.01$ &$0.261\pm0.01$ &$0.261\pm0.01$ &$0.261\pm0.01$ &$0.261\pm0.01$ \\ \hline

\mymodellin &$0.249\pm0.01$    &$0.248\pm0.01$     &$0.248\pm0.01$     &$0.248\pm0.01$   &$0.247\pm0.01$   \\
\mymodelmlp &\underline{$0.289\pm0.01$}    &\underline{$0.283\pm0.02$}     &$\underline{0.283\pm0.01}$     &$\underline{0.280\pm0.01}$   &$0.288\pm0.01$ \\
\mymodelMlp &$0.270\pm 0.02$ &$0.261\pm0.01$    &$0.255\pm0.01$     &$0.254\pm 0.01$    &$0.267\pm 0.02$ \\
\hline
\end{tabular}
\label{tab:syn_exres_feat_full}
\end{table}

\begin{table}[!h]
\setlength{\tabcolsep}{3pt}
\caption{Average Precision (mean) with standard deviation on the synthetic datasets with varying levels of label homophily (see definition \ref{def:homophily}). $r_{homo}$ refers to the fraction of the varying label homophily of the graph with all true features. It was not possible to run \lspe on our synthetic datasets due to its large runtime.}
\centering
\small
\begin{tabular}{l|ccccc}
\hline
\multirow{2}{*}{Method}     &\multicolumn{5}{c}{$r_{homo\_clean}$}\\
    & $0.2$               &$0.4$                &$0.6$                &$0.8$               &$1.0$\\
\hline 
\rule{0pt}{2.5ex}\mlp        &$\underline{0.362\pm0.00}$ &$0.354\pm0.00$ &$0.355\pm0.01$ &$0.360\pm0.01$ &$0.355\pm0.01$\\ 
\deepwalk                   & $0.181\pm0.71$  &$0.522\pm1.41$   &$\underline{0.813\pm0.95}$   &$0.869\pm1.92$ &$0.552\pm0.67$\\ 
\majorityvote               &$\mathbf{0.477\pm0.02}$ &$\mathbf{0.926\pm0.01}$ &$\mathbf{0.957\pm0.01}$ &$\mathbf{0.938\pm0.01}$ &$\mathbf{0.847\pm0.02}$\\\hline
\lanc                       &$0.179\pm0.01$ &$0.397\pm0.02$ &$0.473\pm0.03$ &$0.500\pm0.03$ &$0.634\pm0.05$\\
\gcn                        &$0.261\pm0.01$ &$0.308\pm0.01$ &$0.360\pm0.00$ &$0.430\pm0.01$ &$0.450\pm0.01$\\ 
\gat                &$0.196\pm0.01$ &$0.341\pm0.01$ &$0.353\pm0.00$ &$0.399\pm0.02$ &$0.403\pm0.01$ \\ 
\graphsage  &$0.317\pm0.01$ &$0.408\pm0.00$ &$0.437\pm0.00$ &$0.484\pm0.00$ &$0.491\pm0.01$ \\ 
\hgcn      &$0.300\pm0.01$ &$0.487\pm0.01$ &$0.519\pm0.01$ &$0.617\pm0.03$ &$0.618\pm0.01$\\
\fsgnn &$0.240\pm0.00$    &$0.433\pm0.03$   &$0.419\pm0.01$   &$0.328\pm0.07$   &$0.371\pm0.08$ \\
\gcnlpa     &$0.174\pm0.01$ &$0.277\pm0.00$ &$0.352\pm0.01$ &$0.414\pm0.00$ &$0.346\pm0.01$\\ 
\idgnn  &$0.261\pm0.01$ &$0.308\pm0.01$ &$0.363\pm0.00$ &$0.432\pm0.01$ &$0.457\pm0.01$\\ \hline

\mymodellin   &$0.256\pm0.01$     &$0.468\pm0.01$    &$0.687\pm0.03$    &$0.812\pm0.02$    &\textbf{$0.445\pm0.02$}\\
\mymodelmlp  &\textbf{$0.288\pm0.01$} &$0.492\pm0.01$    &$0.701\pm0.00$    &$0.910\pm0.02$    &\underline{$0.832\pm0.03$}\\
\mymodelMlp     &$0.269\pm0.03$    &\underline{$0.527\pm0.01$}    &$0.795\pm0.02$ &\underline{$0.916\pm0.01$}   &$0.811\pm0.04$\\
\hline
\end{tabular}
\label{tab:syn_exres_homo_full}
\end{table}



\subsection{Broader Impact} 
\label{app:broader_impact}
In the progress analysis of GNNs multi-label node classification has been ignored. We fill in this gap and hope to encourage the community to look into this more realistic and general case of node classification. Our approach can be useful in various applications having positive societal impact like protein function prediction using the multi-labelled protein interaction network. As our method usually more diverse information about the node, there could be possible implications for the privacy of the training data which remain to be analysed.

\end{document}